\RequirePackage{fix-cm}
\documentclass[twocolumn]{svjour3}          %
\smartqed  %
\usepackage{amsfonts}
\usepackage{amssymb}
\usepackage{amsmath}
\usepackage{multirow}
\usepackage{bbding}
\usepackage{graphicx}
\usepackage[numbers]{natbib}
\usepackage{bm}
\usepackage{multirow}
\usepackage{lscape}
\usepackage{booktabs}
\usepackage[%
  colorlinks,bookmarksopen,bookmarksnumbered,citecolor=red,urlcolor=red
]{hyperref}

\begin{document}
\title{Single Person Pose Estimation: A Survey}

\author{Feng Zhang        \and
        Xiatian Zhu        \and
        Chen Wang   %
}

\institute{Feng Zhang \at
              School of Computer Science and Technology, Nanjing University of Posts and Telecommunications, Nanjing, China.
              \email{zhangfeng.ac@outlook.com}  \\
          Xiatian Zhu \at 
              Centre for Vision, Speech and Signal Processing, University of Surrey, Guildford GU2 7XH, United Kingdom.
              \email{eddy.zhuxt@gmail.com} \\
          Chen Wang \at 
              School of Computer Science and Engineering, University of Electronic Science and Technology of China, Chengdu, China.
              \email{wangchen199179@gmail.com} \\
}

\maketitle

\begin{abstract}
Human pose estimation in unconstrained images and videos is a fundamental computer vision task.
To illustrate the evolutionary path in technique, in this survey we summarize representative human pose methods in a structured taxonomy, with a particular focus on deep learning models and single-person image setting.
Specifically, we examine and survey all the components of a typical human pose estimation pipeline, including {\em data augmentation},
{\em model architecture and backbone},
{\em supervision representation},
{\em post-processing},
{\em standard datasets},
{\em evaluation metrics}.
To envisage the future directions, 
we finally discuss the key unsolved problems and potential trends for human pose estimation.

\end{abstract}

\section{Introduction}
Human Pose Estimation (HPE) is a fundamental problem in computer vision. 
HPE aims to obtain the spatial coordinates of human body joints in a person image, with a wide variety of applications such as action recognition \cite{zhang2019view}, person re-identification \cite{zheng2019pose}, semantic segmentation \cite{liang2018look}, 
human-robot interaction \cite{xu2020multi}, etc.
It is challenging due to heavy occlusions, varying clothing styles, poor lighting conditions and various shooting angles.
Earlier HPE methods  \cite{fischler1973representation,felzenszwalb2005pictorial,andriluka2009pictorial,yang2011articulated,johnson2011learning} adopt hand-crafted features to describe human body.
Mostly, these methods aims to learn the underlying relations between different body parts, e.g., using the seminal pictorial structure model \cite{fischler1973representation}.
However, they are limited in accuracy especially under severe occlusions and complex lighting conditions, partly due to
less expressive representations.

In the past years, inspired by the great success of deep learning 
in image recognition \cite{krizhevsky2012imagenet} characterized by end-to-end feature learning capability, researchers have introduced 
an increasing number of deep human pose models
\cite{toshev2014deeppose,wei2016convolutional,newell2016stacked,xiao2018simple,sun2019deep} and continuously refresh the state-of-the-art model performance on standard data benchmarks.
Deep learning based human pose estimation has become
a more exiting and active research area since 
the first introduction of deep learning in 2014 \cite{toshev2014deeppose}.
Given the large body of deep human pose estimation works
published thus far, we consider that it is time to 
review this field for three main purposes:
(1) To reflect its development trajectory and process by extensively
surveying previous methods and algorithms;
(2) To structure different classes of human pose methods;
(3) To provide a holistic picture and schematic overview
of the whole field.
These are not only useful and informative for the beginners to understand quickly
the current state-of-the-art methods for human pose estimation,
but also indicative and inspiring for the experts 
to further push the forefront edges.

There exist a number of surveys on human pose estimation in the literature \cite{ji2009advances,liu2015survey,gong2016human,sarafianos20163d,wang2018rgb,dang2019deep,chen2020monocular}.
However, most of existing surveys \cite{sarafianos20163d,wang2018rgb,ji2009advances,dang2019deep} focus on basic foundations and knowledge, 
without comprehensive review on deep learning human pose estimation methods. 
Whilst \cite{liu2015survey,gong2016human} focus on reviewing human pose methods, they are largely out-of-date with a high desire for covering the latest research advances.
This survey timely solves this need with focus on 
deep learning based human pose estimation on single-person images in particular. 
This is because multi-person human pose estimation can be easily decomposed into person detection and pose estimation,
which represents the state-of-the-art solutions \cite{toshev2014deeppose,wei2016convolutional,newell2016stacked,xiao2018simple,sun2019deep}.

\subsection{Overview}
\begin{figure*}
    \centering
    \includegraphics[width=1.0\textwidth]{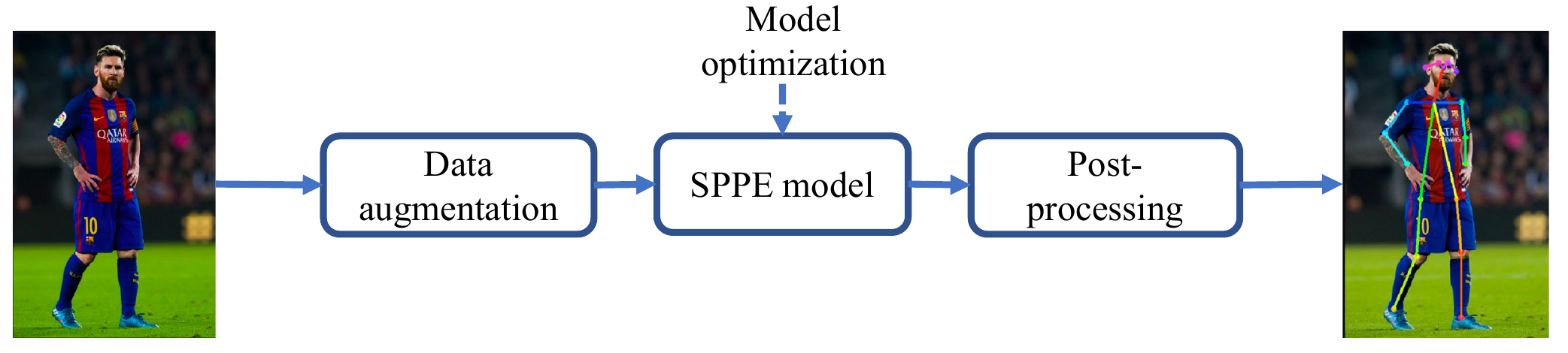}
    \caption{The deep learning pipeline of human pose estimation in the single-person setting.}
    \label{fig:pipeline}
\end{figure*}

To organize a large number of existing deep learning HPE
methods in an intuitive manner, we propose a novel taxonomy 
from the model component perspective.
Table \ref{tab:summary} show our proposed taxonomy with a system pipeline 
illustrated in Fig. \ref{fig:pipeline}.
Accordingly, our survey is organized as the following:
Section \ref{dataprocessing} describes the common data augmentation strategies that are important for boosting the model performance.
Section \ref{model} presents typical model architectures and network backbones.
Section \ref{postprocessing} reviews the largely ignored yet significant post-processing.
Section \ref{learning} details the supervision representation used in optimization.
Section \ref{data} and Section \ref{metrics} summarize the standard human pose datasets and performance evaluation metrics, respectively.

\begin{table*}[]
\caption{The taxonomy of human pose estimation in the model component perspective.
}
\label{tab:summary}
\bgroup
\def\arraystretch{1.9}%
\resizebox{\textwidth}{!}{%
\begin{tabular}{c|c|c}
\toprule
\bf Perspective &\bf Category &\bf Sub-category \\ \hline \hline

\multirow{4}{*}{Data augmentation} & \multirow{2}{*}{Random data augmentation} & Generic data augmentation \cite{tompson2014joint,pfister2015flowing} \\ \cline{3-3} 
 &  & Pose-specific data augmentation \cite{ke2018multi,wang2018mscoco,park2020data} \\ \cline{2-3} 
 & \multirow{2}{*}{Optimized data augmentation} & Neural Architecture Search (NAS) based methods \cite{hou2020augmented} \\ \cline{3-3} 
 &  & Generative Adversarial Network (GAN) based methods \cite{peng2018jointly} \\ 
 \hline
 
\multirow{9}{*}{Deep human pose model} & \multirow{5}{*}{Model architecture} & Sequential architectures \cite{jain2013learning,fan2015combining,pfister2015flowing,rafi2016efficient,nie2018human,sun2019deep,tang2019does,su2019multi} \\ \cline{3-3} 
 &  & Cascaded architectures \cite{toshev2014deeppose,tompson2015efficient,wei2016convolutional,sun2017human,newell2016stacked,yang2017learning,ke2018multi,chen2018cascaded,liu2018cascaded,ning2017knowledge,zhang2019human} \\ \cline{3-3} 
 &  & Recurrent architectures \cite{pfister2015flowing,belagiannis2017recurrent,carreira2016human,gkioxari2016chained} \\ \cline{3-3} 
 &  & Adversarial architectures \cite{chen2017adversarial,chou2018self} \\ \cline{3-3} 
 &  & Neural Architecture Search (NAS)  \cite{zoph2016neural,baker2016designing,liu2018darts,yang2019pose,gong2020autopose} \\ \cline{2-3} 
 & \multirow{4}{*}{Network design} & Multi-scale feature learning  \cite{tompson2015efficient,newell2016stacked,yang2017learning,ning2017knowledge,liu2018cascaded,chen2018cascaded,cao2019anti,sun2019deep,su2019multi,artacho2020unipose} \\ \cline{3-3} 
 &  & Prior knowledge incorporation \cite{ning2017knowledge,bulat2016human,xia2017joint,nie2018human} \\ \cline{3-3} 
 &  & Spatial relation modelling \cite{chen2014articulated,tompson2014joint,yang2016end,chu2016crf,chu2016structured,sun2017compositional,lifshitz2016human,chen2017adversarial,tang2019does,zhang2019human} \\ \cline{3-3} 
 &  & Attention mechanism \cite{chu2017multi,liu2018cascaded,su2019multi} \\ \hline

\multirow{4}{*}{Supervision representation} & \multirow{2}{*}{Heatmap based methods} & Heatmap regression \cite{tompson2014joint,pfister2015flowing,wei2016convolutional,liu2018cascaded,sun2019deep} \\ \cline{3-3} 
 &  & Heatmap regression and offset regression \cite{papandreou2017towards}  \\ \cline{2-3} 
 & \multirow{2}{*}{Coordinate based methods} & Direct coordinate regression \cite{toshev2014deeppose,fan2015combining,sun2017human} \\ \cline{3-3} 
 &  & Coordinate regression through heatmap \cite{sun2018integral} \\ 
 \hline
 
 \multirow{2}{*}{Post-processing} 
 & Heuristic methods & Empirical design \cite{newell2016stacked} \\ \cline{2-3} 
 & Data-driven methods & 
 Distribution learning \cite{zhang2020distribution,yang2020train}
 \\

 \bottomrule
\end{tabular}%
}
\egroup
\end{table*}

\section{Data Augmentation}
\label{dataprocessing}

As a generic and common model training strategy, data augmentation
is critical for the generalization performance of deep learning methods
including human pose models,
due to the data hungry nature of deep neural networks \cite{cirecsan2010deep,krizhevsky2012imagenet, simonyan2014very}
and the difficulty of collecting large training data.
Often it is considered as a regularization 
which can mitigate the notorious overfitting problem.
In general, data augmentation would generate more training samples 
by introducing some modification to the original training data
whilst keeping the same semantic labels.
There are random and optimized data augmentation for training human pose models,
as detailed below.
\subsection{Random Data Augmentation}
Data augmentation operations are often randomly applied
to a training sample during training.
The common augmentation operations include flipping, rotating, scaling, occluding, and color jittering \cite{pfister2014deep}.

\paragraph{\bf Flipping.} 
Flipping can improve the model's robustness against imaging direction.
There are two types of flipping: horizontal flipping and vertical flipping.
Horizontal flipping is used more than the vertical one.
In model training, the flipping augmentation is usually triggered at a certain probability, e.g. 50\%.

\paragraph{\bf Rotation.} 
Human bodies may be lying and leaning with the body spines exhibiting different angles in imbalanced distribution.
This imbalance may impose negative effect to model training.
Rotation augmentation is a simple solution for mitigating this.
It is performed by rotating the image clockwise or counterclockwise around the center of human body.
Typically, the rotation angle ranges from -30 to 30 degrees.
As flipping, the rotation operation is also triggered at a preset probability.

\paragraph{\bf Scaling.} 
Body scale variation is often large 
in unconstrained images which might negatively affect the performance of human pose estimation methods.
To alleviate this issue, scaling augmentation is adopted to improve the model robustness for image scales.
In training, a random scaling factor is selected and applied. 

\paragraph{\bf Color jittering.}
Human person images are typically collected in a variety of environments with different lighting conditions.
To make the model suitable to cope with such image data, color jittering is effective to simulate diverse lighting conditions in training. 

\paragraph{\bf Occlusion.
}
Occlusion is an inevitable challenge 
due to uncontrolled shooting angle and complex scene.
Excepted collecting samples which is largely limited, synthesizing images with occlusions turns out to be cheaper and easier for improving model training.

\paragraph{\bf Pose-specific augmentations.
}
All the above augmentation operations are 
generally applicable for many different computer vision tasks.
Besides, pose-specific augmentation techniques \cite{ke2018multi,wang2018mscoco,park2020data} have been proposed additionally.
Specifically, in the real world, 
it is often the case that human body joints are obscured by other objects;
In crowded scenes, there may be multiple joints of the same class that can greatly confuse a model.
Under such observation, Ke et al. \cite{ke2018multi} proposed a keypoint-masking augmentation method
to synthesize hard training samples with self-occlusion and mutual interference.
Considering that there are a large number of samples with a certain part of human body and the upper body presents easier-to-detect joints than the lower body,
Wang et al. \cite{wang2018mscoco} presented a half-body augmentation strategy for 
more specific operation.
It is observed that samples with few keypoints in the training set are corner cases
often simply ignored by existing methods.
To better solve this problem, Park et al. \cite{park2020data} introduced a body-part-cropping augmentation approach by randomly selecting and cropping a proportion of human body at different scales to increase the diversity of training data.

\subsection{Optimized Data Augmentation}
Random data augmentation is simple yet less efficient.
This can be addressed by learning to augment training data,
i.e., optimized data augmentation.
It is a relatively new research direction in human pose estimation. 
Two representative types of learning framework in the literature are 
{\em adversarial learning}
and
{\em neural architecture search}.

The first type usually employs the generative adversarial networks (GANs).
The key idea is to generate {\em hard training samples}
that can more effectively incur training loss and therefore expedite model learning.
For example, Peng et al. \cite{peng2018jointly} proposed Online Adversarial Augmentation Network (OAAN) jointly learned with an existing HPE model. 
OAAN is designed to generate hard adversarial samples with a reward-penalty policy in a way that the target HPE model can improve its robustness against hard samples.
Bin et al. \cite{bin2020adversarial} proposed a Semantic Data Augmentation (SDA) method.
It aims to solve the limitation of random data augmentation \cite{ke2018multi} that generated samples by a naive copying-and-pasting body part strategy have limited diversity and unrealistic appearance.
SDA involves constructing semantic body part pool for generation of hard training samples.
Specifically, a semantic body part pool is first constructed by a human parsing algorithm; Then, hard training samples are generated with an optimal pasting configuration in an adversarial learning manner.

Unlike adversarial learning based methods,
the second type instead aims to optimize the data augmentation application process.
For instance, Hou et al. \cite{hou2020augmented} first defined a search space with different operators such as flipping, cropping, rotation.
By treating the sequence of applying these operators as a trainable component,
they turned this problem into a discrete optimization problem and solved it via an existing differentiable search algorithm \cite{hanxiao2019}.

\section{Model Architecture and Network Design}
\label{model}

According to the theory that the structure of neural networks determines its function,
the human brain has highly structured connections at birth 
which determines what functions it has. 
These functions in the brain are enhanced through learning subsequently.
Like these networks in the brain, their architectures and backbones 
play a very important role in HPE. 
How to design a good architecture and backbone encoding human prior knowledge remains an open problem.
The models taking the scale variability and severe occlusions into consideration make it possible to learn the complex mapping from natural image to body pose.
This section will describe the recent researches of the HPE model from the two perspectives: the architecture and backbone.

\subsection{Model Architecture}

The architecture of HPE model can be classified into four categories: sequential, cascaded, recurrent and adversarial architectures.

\paragraph{\bf (I) Sequential architecture.
} 
Typical sequential models \cite{jain2013learning,fan2015combining,pfister2015flowing,rafi2016efficient,nie2018human,sun2019deep,tang2019does,su2019multi}
construct a simple yet intuitive framework by
a series of basic layers such as convolution,
pooling, fully connection to learn a mapping from input images to joint coordinates.
Jain et al. \cite{jain2013learning} proposed a CNN architecture with fully connected layers. 
The network slides over the input image with an overlapped sliding window to detect the presence of joints.
Fan et al. \cite{fan2015combining} developed a dual-source network architecture that uses the raw image with a holistic view and the image patches with a local view as model inputs to improve the accuracy of HPE.

To avoid kinematically impossible predictions, Pfister et al. \cite{pfister2015flowing} presented a novel architecture that includes a spatial fusion layer to learn spatial dependencies of joints implicitly.
Rafi et al. \cite{rafi2016efficient} adopted an efficient multi-scale network architecture based on the Inception network
to reduce the model complexity and improve the model's ability to perceive larger context.
Nie et al. \cite{nie2018human} proposed the Parsing Induced Learner (PIL) method to 
facilitate the HPE model with parsing information.
Sun et al. \cite{sun2019deep} proposed a network architecture that preserves high-resolution features and fused low-resolution features with high-resolution features to 
learn robust multi-scale feature representations.

Previous HPE models learn the shared feature representations for all body parts,
Tang et al. \cite{tang2019does} found that the shared feature representations
are beneficial to related parts but fail to improve the learning of other parts.
To address this problem, they obtained the correlations between joints through statistical analysis and designed a sequential network architecture with multiple branches to incorporate structured knowledge of human body into the network architecture design.
Su et al. \cite{su2019multi} designed the Channel Shuffle Module (CSM) and the Spatial Channel-wise Attention Residual Bottleneck (SCARB) to introduce the attention mechanism into the networks.
Specifically, the CSM can improve the information exchange across channels and the SCARB is able to boost the residual learning. %
Inspired by the Atrous Spatial Pyramid Pooling (ASPP) in semantic segmentation,
Artacho et al. \cite{artacho2020unipose} proposed the Waterfall Atrous Spatial Pooling (WASP) module to enlarge the Field of View (FOV) and obtain multi-scale representations.

\paragraph{\bf (II) Cascaded architecture.
} 
The cascaded architecture usually repeats several modules to 
refine the prediction.
Toshev et al. \cite{toshev2014deeppose} presented the first application of CNNs to human pose estimation and designed a cascaded architecture.
The regressor in the first stage aims at estimating coarse body pose
and the subsequent regressors are designed to learn the displacement of the part locations from the previous stage to the ground-truth locations.  %

Traditional architectures for classification use down-sampling layers such as pooling to maintain invariance, reduce computation and improve efficiency, but these pooling layers in the network are not appropriate for the HPE task and may hurt the model performance.  %
Thompson et al. \cite{tompson2015efficient} reduced the number of down-sampling layers and implemented a sliding-window detector with overlapping contexts to produce coarse predictions and then used the Siamese network to refine these coarse locations. 
Wei et al. \cite{wei2016convolutional} designed a multi-stage architecture called Convolutional Pose Machine (CPM) to learn long-range dependencies between joints.
The CPM consists of several stages and each stage operates on the predicted heatmaps from the previous stage. 
The spatial relations of joints are captured through increasing the size of receptive field for each subnetwork in the CPM.
Sun et al. \cite{sun2017human} analyzed the relative locations of joints and found the uniform distribution of these relative locations 
has a negative effect on the learning procedure.
To make the learning procedure easier, they proposed a novel network with normalization module which is composed of body normalization and limb normalization.

To extract multi-scale features and deal with ambiguous prediction, Newell et al. \cite{newell2016stacked} designed the hourglass module by repeating the bottom-up and top-down processing.
Yang et al. \cite{yang2017learning} introduced the fractional pooling into the residual block and proposed the Pyramid Residual Module (PRM) to
learn scale-invariant representations from training data.
Ke et al. \cite{ke2018multi} proposed a multi-scale structure learning framework based on the revised hourglass module and utilized additional multi-scale regression to perform global optimization of structure configuration.
There are occluded joints, invisible joints, and very complex backgrounds in the real world.
To cope with these challenging cases, Chen et al. \cite{chen2018cascaded} proposed a novel network structure called Cascaded Pyramid Network (CPN) consisting of a GlobalNet and a RefineNet.
The GlobalNet is responsible for the easy samples and the RefineNet aims at dealing with those challenging samples.
Liu et al. \cite{liu2018cascaded} designed a Cascaded Inception of Inception Network (CIOIN).
The Inception of Inception (IOI) block in CIOIN is able to fuse features with different semantics and preserve the scale diversity in them.
The Attention-modulated IOI (AIOI) block is capable of 
adjusting the importance of features according to the context.
The CIOIN architecture employs a multiple iterative stage to capture long-range dependencies.
Ning et al. \cite{ning2017knowledge} constructed a fractal network by using the hourglass modules and Inception-ResNet blocks to regress human pose without explicit graphical modelling.
To extract contextual information, 
Zhang et al. \cite{zhang2019human}
proposed the Cascade Prediction Fusion (CPF) to fuse predictions and features from the previous stage.
To learn relations between joints,
they further designed the Pose Graph Neural Network (PGNN) to model these relations according to the body structure.

\paragraph{\bf (III) Recurrent architecture.
}
Feedforward architectures are able to learn hierarchical feature representations from data,
but they fail to capture the connections between joints and the associations between features from different layers.  %
Recurrent architectures enhance the feature representations and incorporate top-down feedback into the feedforward architectures to complement their feature representations, establishing relations between the input space and the output space.
Pfister et al. \cite{pfister2015flowing} made the first attempt to
model features from different layers by leveraging the proposed spatial fusion layer. 
Inspired by Pfister, Belagiannis et.al \cite{belagiannis2017recurrent}
presented a novel recurrent model including a feed-forward module and several recurrent modules. These recurrent modules can effectively 
revise predictions with iterative feature fusion.
Carreira et al. \cite{carreira2016human} 
implemented a self-correction model which simulates the recurrent mechanism in the human brain to learn a residual displacement from current prediction to the ground truth progressively in an Iterative Error Feedback (IEF) manner.
Unlike previous studies \cite{pfister2015flowing,belagiannis2017recurrent,carreira2016human} on recurrent architecture, 
Gkioxari et al. \cite{gkioxari2016chained}
defined a fixed ordering of joints based on the kinematic tree of the human body and modelled the dependencies between adjacent joints by learning a conditional distribution to predict joints one by one.

\paragraph{\bf (IV) Adversarial architecture.
}
GAN consists of a generator and a discriminator, 
and is trained by playing minimax game between the generator and the discriminator. 
The generator synthesizes samples capable of confusing the discriminator, while the discriminator continuously improves its capability in distinguishing fake samples from the true data distribution. 
Recent applications of GAN in human pose
estimation include \cite{chen2017adversarial} and \cite{chou2018self} view the generator as a HPE network.
They treat the discriminator as a proxy of the loss function to guide the learning process of the generator by imposing geometric constraints and prior knowledge through the discriminator.
To exploit body configurations and enforce structure constraints on HPE model, Chen et al. \cite{chen2017adversarial} developed two discriminators
(a pose discriminator and a confidence discriminator).
The pose discriminator is responsible for identifying whether the locations of joints estimated by the generator are reasonable or not, and the confidence discriminator can utilize the geometric structure of the human body to evaluate these predictions with high confidence, especially the occluded joints with high confidence.
Different from \cite{chen2017adversarial}, Chou et al. \cite{chou2018self} proposed a discriminator to reconstruct heatmaps.
For $j$-th joint, the discriminator is optimized by minimizing the loss $L_{fake}$ between the generated heatmaps $\{\hat{C}_j\}_{j=1}^{J}$ and the reconstructed ones $\{D(\hat{C}_j,X)\}_{j=1}^{J}$ (Eq.\eqref{eq:fake}), and it tries to maximize the loss $L_{real}$ between the ground-truth heatmaps $\{C_j\}_{j=1}^{J}$ and the reconstructed ones $\{D(C_j,X)\}_{j=1}^{J}$ (Eq.\eqref{eq:real}).
\begin{equation}
    \label{eq:real}
    L_{real} = \sum_{j=1}^{M} (C_j - D(C_j,X) )^2
\end{equation}

\begin{equation}
    \label{eq:fake}
    L_{fake} = \sum_{j=1}^{M} (\hat{C}_j - D(\hat{C}_j,X) )^2
\end{equation}

\paragraph{Neural architecture search.
}
In practice, the hand-designed architecture needs to be adjusted according to the requirements of a specific task.
However, these modifications involve tedious verification process.
Therefore, Neural Architecture Search (NAS) is presented to 
find good network architecture with limited computing resources and little human intervention.
Zoph et al. \cite{zoph2016neural} and Baker et al. \cite{baker2016designing} introduced Reinforcement Learning (RL) into NAS by modelling the searching process as a Markov Decision Process (MDP).
However, this kind of methods consume too many computing resources making them infeasible to practical applications.

In order to reduce the search time, Liu et al. \cite{liu2018darts} proposed the DARTs method which implements a continuous relaxed representation of the architecture. 
Specifically, they replaced the discrete non-differentiable search space in RL based approaches to obtain the optimal architecture by gradient descent efficiently.
Motivated by Convolutional Neural Fabrics \cite{saxena2016convolutional} and DARTS \cite{liu2018darts}, 
Yang et al. \cite{yang2019pose} designed the part representation
according to the structure of human skeleton and parameterized
the Cell-based Neural Fabric (CNF) to perform the gradient-based search at the micro and macro search space.
Gong et al. \cite{gong2020autopose} presented a novel NAS framework called AutoPose which 
defines a hierarchical multi-scale search space and utilizes both gradient-based cell-level search and reinforcement learning based network-level search to discover multiple parallel branches of diverse scales.

\subsection{Network Design}

\paragraph{\bf Multi-scale feature learning.
}
The main purpose of the research of HPE is to extract multi-scale features to cope with various scales of the human body and to obtain a larger receptive field for global inference. 
Tompson et al. \cite{tompson2015efficient} proposed a two-stage network
with pose refinement implemented by a Siamese network.
The Siamese network takes the coarse heatmaps with different resolutions as inputs to refine predictions.

To handle body parts with different scales such as the face, hand, foot, etc,
Newell et al. \cite{newell2016stacked} presented the stacked hourglass network formed by stacking several hourglass modules with pooling and upsampling operations to capture features at every scale.
Motivated by the residual learning \cite{he2016deep}, Wei et al. \cite{yang2017learning} proposed several variants of Pyramid Residual Modules (PRMs) simulating the mechanism of human visual perception to capture multi-scale features.
Ning et al. \cite{ning2017knowledge} adopted the design of Inception block and residual block, and proposed the Inception-ResNet module to construct a fractal network structure which captures the interdependence between joints across different scales and resolutions.

To deal with various appearance and occlusion in human pose estimation,
Liu et al. \cite{liu2018cascaded} proposed a Cascaded Inception of Inception Network (CIOIN).
The Inception of Inception (IOI) block in the CIOIN not only is able to preserve scale diversity but also can be upgraded to 
the Attention-modulated IOI (AIOI) block to reweight the features with different scales.
Chen et al. \cite{chen2018cascaded} designed the Cascaded Pyramid Network (CPN) architecture based on the ResNet backbone, and they integrated the U-shape structure into the architecture design to create a feature pyramid that is able to preserve spatial information of multi-scale features.
Cao et al. \cite{cao2019anti} introduced a stronger low-level feature module called Feature Pyramid Stem (FPS).
The FPS module is capable of extracting target-specific features in valid regions, and progressively fusing features with different resolutions to improve the quality of representations.

Most of the existing architectures utilize the high-to-low and low-to-high process to learn the multi-scale feature representations.
However, recovering from low-resolution representations does not bring more useful information.
How to maintain high-resolution and multi-scale representations is a
non-negligible concern in architecture design.
Sun et al. \cite{sun2019deep} presented a novel architecture, namely HighResolution Net (HRNet), which can maintain high-resolution representations through the whole process. 
Su et al. \cite{su2019multi} proposed a Channel Shuffle Module (CSM) to further enhance the cross-channel information exchange between feature maps with different scales.

To exploit the contextual information, enlarge the field-of-view,
and enhance multi-scale representations, 
Artacho et al. \cite{artacho2020unipose} proposed the Waterfall Atrous Spatial Pooling (WASP) module.
The WASP module employs cascaded atrous convolutions at increasing rates in the parallel architecture to obtain larger field-of-view.

\paragraph{\bf Prior knowledge incorporation.
}
Although end-to-end networks can learn discriminative feature representations, 
the HPE model trained from limited data cannot describe the actual distribution.
How to encode prior knowledge into neural networks is a very complicated task. 
To solve this problem, 
Ning et al. \cite{ning2017knowledge} designed two types of external knowledge that includes geometrical features and Histogram of Gradients (HOG) based features.
They utilized the proposed knowledge projection module to decode the abstract external knowledge and injected the decoded knowledge into the fractal network.
Bulat et al. \cite{bulat2016human} proposed a part detection network to 
capture contextual information and structured relations.
The part detection network guides the pose regression network to learn the structure of body parts, improving the predictions of occluded parts with low confidence scores.

Motivated by the correlations between part segmentation and pose estimation, Xia et al. \cite{xia2017joint} proposed a prior knowledge 
incorporation schema to exploit segmentation results to improve HPE.
Specifically, semantic part prior knowledge from part Fully Convolutional Network (FCN) is fed to Fully-connected Conditional Random Field (FCRF) to avoid unreasonable predictions from the pose FCN.
Nie et al. \cite{nie2018human} proposed a parsing-guided HPE method which uses the Parsing Induced Learner (PIL) to learn the parameters of adaptive convolution from the features of the parsing encoder.
Specifically, they leverages the adaptive convolution to adjust features from the pose encoder to obtain accurate predictions.

\paragraph{\bf Spatial relation modelling.
}
The dimension of input space is determined by many factors such as deformation caused by the varying viewing angles, changes in lighting and clothing, and occlusion induced by human-object interaction.
The mapping from image to pose is highly non-linear, which makes it 
necessary to exploit the spatial inter-dependencies between joints to 
improve the representation ability of HPE model. 
Chen et al. \cite{chen2014articulated} defined a tree-structure graphic model to learn spatial relations by conditional probabilities from unary term and Image Dependent Pairwise Relationships (IDPR) term, and adopted Dynamic Programming (DP) algorithm to find the optimal configuration for person instances. 
Tompson et al. \cite{tompson2014joint} proposed a Markov Random Field (MRF) model which contains
a unary term for modelling joint locations and a pairwise term for modelling the conditional distribution of neighbouring joints.
They implemented the MRF model by the message passing network to eliminate incorrect predictions.

Different from the MRF-liked model \cite{tompson2014joint},
Yang et al. \cite{yang2016end} blended the appearance mixtures into the loopy model to
enforce deformation constraints, which allows the HPE model to handle large pose variations.
Instead of modelling structures by graphical models \cite{tompson2014joint,yang2016end,chu2016crf},
Chu et al. \cite{chu2016crf} proposed a CRF-CNN framework which extends the loopy model by adding a novel pairwise term to learn the relations and structures among features.
To refine feature maps, Chu et al. \cite{chu2016structured} presented a bi-directional structured feature learning model by utilizing geometrical transform kernels to pass on information between feature maps for each joint.
Sun et al. \cite{sun2017compositional} introduced an intuitive yet stable bone-based representation which is easier to learn and more stable than traditional joint-based representation to encode the inter-dependencies inside the structure of body pose.
Lifshitz et al. \cite{lifshitz2016human} designed a voting representation 
by log-polar binning and then aggregated voting maps using a large deconvolution kernel. They added a novel image-based consensus voting binary term to remove the independence assumption in traditional binary term by averaging over all locations in the voting maps. 

Motivated by the GAN, Chen et al. \cite{chen2017adversarial} proposed a pose discriminator which implicitly exploits the spatial inter-relations of joints to distinguish the fake poses from the real ones.
Tang et al. \cite{tang2019does} analyzed the correlations between different body parts, and obtained related parts by applying spectral clustering to the correlation matrix.
They designed a part-based branching network according to the compositionality of human bodies to learn specific high-level features for better performance.

Graphical models involving multiple sequential updates usually incur accumulative error. 
Instead of defining a explicit graphical model to learn the structure of the human body, Zhang et al. \cite{zhang2019human} integrated Pose Graph Neural Network (PGNN) into the HPE network to exploit spatial contextual information from neighboring joints.

\paragraph{\bf Attention mechanism.
}
Attention mechanism in the human visual system helps the brain ignore irrelevant regions and pay more attention to important information. 
In the field of HPE, attention mechanism can also be integrated into existing networks to help HPE models focus on relevant areas.
Chu et al. \cite{chu2017multi} designed three types of attention modules for HPE.
The multi-resolution attention within the hourglass module aims to refine features,
and the multi-semantics attention across several stacks of hourglass is used to capture various semantic information.
Besides, the hierarchical attention is adopted to encode holistic and local body structure.
Liu et al. \cite{liu2018cascaded} introduced an Inception of Inception (IOI) module to preserve low-level features.
To improve the robustness of the HPE model against scale variety, they incorporated the attention mechanism into the IOI module to form an Attention-modulated IOI (AIOI) module to fuse multi-scale features dynamically.
To learn the relations between the low-level and high-level feature maps,
Su et al. \cite{su2019multi} proposed a Channel Shuffle Module (CSM) implemented by channel shuffle operation to exchange information within feature pyramid, and designed a Spatial Channel-wise Attention Residual Bottleneck (SCARB) to highlight the spatial and channel-wise information from regions of interest.
\section{Post-processing}
\label{postprocessing}

A fundamental challenge in human pose estimation is that, the gap between {\em continuous joint space} and {\em discrete heatmap space} results in quantization error.
Despite it is important for model performance,
this error issue is largely ignored with little attention and efforts paid in most studies.
To mitigate this problem, Newell et al. \cite{newell2016stacked} presented a simple, heuristic method. Specifically, they shifted empirically the highest location $\bm{m}$ in heatmap towards the second highest peak $\bm{s}$ by a quarter of pixel to obtain the final output as:
\begin{equation}
	\bm{p} = \bm{m} + 0.25 \frac{\bm{s}-\bm{m}}{\|\bm{s}-\bm{m}\|_2}
	\label{eq:standard_shfit}
\end{equation}
Since then, this ad-hoc method has been extensively used in the follow-up HPE models as {\em de-facto standard} post-processing.
However, the rationale of this method is largely unknown.
Importantly, how does this step affect the final model performance is under-studied.

To solve these issues, Zhang et al. \cite{zhang2020distribution}
conducted a systematic investigation on
post-processing.
They revealed for the first time that
this quantization error could introduce significant
model performance degradation, particularly 
when low-resolution input images are used.
Moreover,
a superior distribution-aware post-processing method based on Taylor series approximation to Gaussian distribution
was invented without re-training already-optimized HPE models.
In particular, by approximating the heatmap distribution with Taylor series, it decodes human joint coordinates as:
\begin{equation}
	\bm{p}
	= \bm{m} - \big(\mathcal{D}''(\bm{m}) \big)^{-1} \mathcal{D}'(\bm{m})
	\label{eq:final_joint}
\end{equation}
\noindent where the first- $\mathcal{D}'(\bm{m})$ and second-gradient $\mathcal{D}''(\bm{m})$ can be estimated efficiently around the outputted peak $\bm{m}$ point only.

Subsequently, treating the predicted heatmap as a composite function of signal and noise, Yang et al. \cite{yang2020train} 
proposed a Distribution-Aware and Error-Compensation (DAEC) decoding method to obtain coordinates of keypoints by integrating over the entire heatmap. 
To achieve more accurate predictions,
Fieraru et al. \cite{fieraru2018learning} proposed a data synthesis technique by simulating the typical erroneous samples and learning a pose refinement model from these synthesized samples to correct the predictions.
Moon et al. \cite{moon2019posefix} proposed a model-agnostic refinement method
that adopts the diagnosis method \cite{ruggero2017benchmarking} to analyze the common localization errors in HPE and utilizes the error statistics to synthesize new training samples.
It is shown that a coarse-to-fine estimation network trained with these samples can improve %
effectively.

\section{Supervision Representation}
\label{learning}
From the supervision representation perspective, 
HPE methods can be classified into 
{\em coordinate-based} \cite{toshev2014deeppose,fan2015combining,sun2017human} and 
{\em heatmap-based} methods \cite{tompson2014joint,pfister2015flowing,wei2016convolutional,liu2018cascaded,sun2019deep}.
Specifically, coordinate-based methods predict/regress the joint coordinates directly.
Direct coordinate regression is intuitive and simple in model design.
However, this approach often reaches only unsatisfactory model performance, largely due to 
that this regression learning task is extremely difficult particularly given unconstrained images.

On the other hand, the heatmap-based methods  \cite{tompson2014joint,pfister2015flowing,wei2016convolutional,liu2018cascaded,sun2019deep} 
take an extra advantage of the spatial structure information in the image to describe the relations between body parts and provide contextual and structural support for model learning.
Usually these methods adopt discrete heatmaps to represent continuous distribution. 
The conversion from discrete heatmaps to coordinates is hence necessary which introduces quantization errors.
To alleviate this problem, Sun et al. \cite{sun2018integral} treated the heatmap as an intermediate representation and proposed a differentiable integral approach to obtain the joint coordinate from the heatmap.
Similar to \cite{sun2018integral}, Nibali et al. \cite{nibali2018numerical} proposed a Differentiable Spatial to Numerical Transform (DSNT) method, as well as employed the Jensen-Shannon divergence to enforce a shape constraint on the distribution of heatmap. %
Besides, Earth Mover’s Distance (EMD) \cite{martin2017wasserstein,liu2020detection} can also be utilized to measure the distribution between ground-truth and predicted heatmaps.
Papandreou et al. \cite{papandreou2017towards} and Zhang et al. \cite{zhang2019exploiting} 
explored an alternative to reduce the quantization error. 
They divided the localization problem into two parts: a detection task and a regression task.
The detection task learns binary heatmaps indicating the regions of body parts, whilst the regression task learns the fractional part of numerical joint coordinates.
It is therefore a coarse-to-fine learning pipeline.
To addressing significant model performance degradation from low-resolution, Wang et al. \cite{wang2021lowres} introduced a novel Confidence-Aware Learning (CAL) method by addressing the fundamental training-testing discrepancy of previous offset learning methods \cite{papandreou2017towards,zhang2019exploiting}.

\section{Human Pose Datasets}
\label{data}
\begin{figure*}
    \centering
    \resizebox{\linewidth}{!}{
        \includegraphics{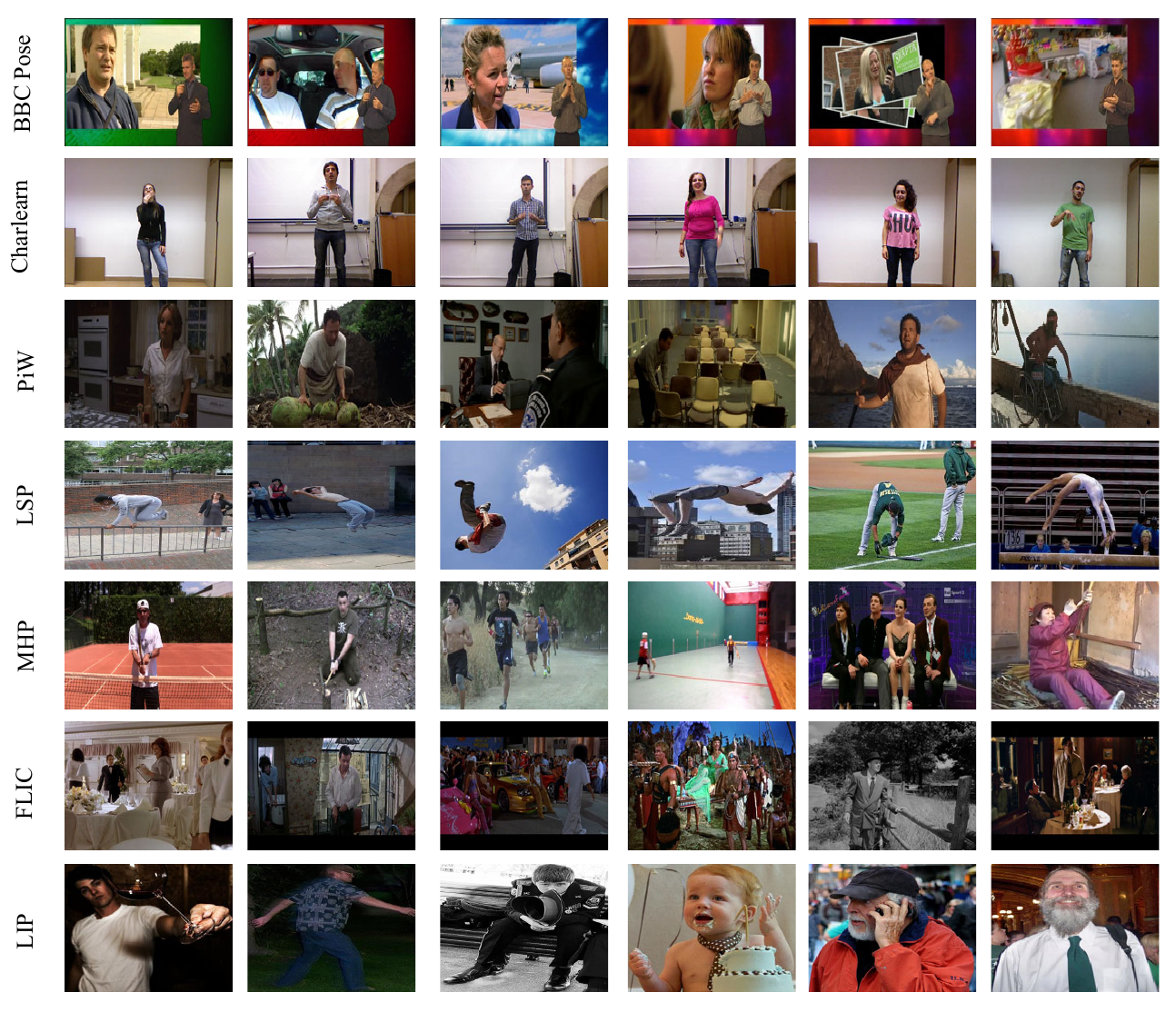}
    }
    \caption{Dataset samples. We show six random image samples for each dataset. 
    }
    \label{fig:samples}
\end{figure*}

This section presents seven popular HPE datasets including BBC Pose datasets, ChaLearn dataset, Poses in the Wild dataset, Frames Labeled In Cinema datasets, Leeds Sports Pose datasets, MPII Human Pose dataset and
Look Into Person dataset. Samples of each dataset are visualized in Fig.\ref{fig:samples}.

\paragraph{\bf BBC Pose dataset.
} 
The BBC Pose dataset contains 20 videos extracted from BBC programmes, each of which is approximately 0.5h to 1.5h in length.
There are 10/5/5 videos in the training set, validation set and test set respectively.
There are 5 signers in the training and validation sets respectively, and 4 signers in the test set. 
Splitting the dataset this way aims to evaluate the generalization performance of HPE model in dealing with different signers. 
The annotation contains 7 joints, i.e., head, shoulders, elbows, wrists.
The labels of training and validation sets are obtained by \cite{buehler2011upper}, while the samples in the test set are manually annotated.
An extended version of this dataset, the extended BBC Pose dataset, includes additional 72 training videos annotated by the tracker of Charles et al. \cite{charles2014automatic}.
In addition, Charles et al. \cite{charles2014automatic} also presented the Short BBC Pose dataset. 
This dataset contains 5 videos with different sleeve lengths compared to the BBC Pose and the extended BBC Pose datasets whose samples contain only long-sleeved signers. 
In order to evaluate the performance of algorithms on different videos accurately,
each video contains 200 hand-annotated test frames and the final test set is composed of 5 videos.

\paragraph{\bf ChaLearn dataset.
}
The Chalearn dataset \cite{escalera2013multi} is a multimodal dataset collected with the Kinect camera, including sound, 2D and 3D poses, RGB and depth images, and segmentation masks indicating regions of the human body from 27 persons.
The dataset is split into 393/287/276 video clips for training, validation and testing, respectively. 
The whole human skeleton includes 20 joints, i.e., hip centre, spine, shoulder centre, head, shoulders, elbows, wrists, hands, hips, knees, feet, ankles. 
The dramatic changes in clothing across videos cause the main challenge for HPE.

\paragraph{\bf Poses in the Wild (PiW) dataset. 
} 
To evaluate the performance of models in the wild, Cherian et al. 
\cite{cherian2014mixing} proposed the Poses in the Wild (PiW) dataset which consists of 30 sequences from 3 Hollywood movies.
The PiW dataset defines 8 human upper-body keypoints (neck, shoulders, elbows, wrists, and mid-torso) and is
characterized by occlusions, severe camera motion and background clutter.

\paragraph{\bf Frames Labeled In Cinema (FLIC) dataset.
} 
Similar to the PiW dataset, Frames Labeled In Cinema (FLIC) dataset \cite{sapp2013modec} is extracted from 30 popular Hollywood movies and annotated by the crowdsourcing marketplace Amazon Mechanical Turk (AMT).
The dataset has a total of 5003 images and is split into two parts: the training set ($\sim$80\%, 3987 images) and the test set ($\sim$20\%, 1016 images).
Different from the PiW dataset with 8 keypoints, the annotation in FLIC dataset has 11 upper-body joints, namely nose, eyes, shoulders, elbows, wrists, and hips.
The FLIC dataset has a full set version where all frames from movies are annotated by the AMT labelling service.

\paragraph{\bf Leeds Sports Pose (LSP) dataset.
} 
In sports activities, there are many challenging cases such as motion blur, severe occlusion, extreme body deformation.
The Leeds Sports Pose dataset \cite{johnson2010clustered} (LSP) is collected from Flickr to solve the pose estimation in sports activities.
The dataset contains 2000 various sports images, 
with a half used for training and the other half for testing. 
Different from PiW and FLIC, LSP pays attention to the full-body pose estimation and defines the human pose with 14 joints, i.e., ankles, knees, hips, wrists, elbows, shoulders, neck, and head top.

\paragraph{\bf MPII Human Pose (MHP) dataset.
}
Deep learning based HPE methods require large amounts of data, 
but the sizes of early datasets are often limited.
Small datasets cannot cover many practical cases and
the lack of diversity could cause model overfitting 
particularly in deep learning.
To address these issues, Andriluka et al. \cite{andriluka20142d} established a taxonomy of daily activities and collected human images from various human activities in YouTube videos to build a large benchmark called MPII Human Pose (MHP) dataset.
Compared with previous datasets, MHP dataset is characterized by its large scale and high diversity.
The MHP dataset offers richer pose annotations including coordinates of 16 joints (i.e., shoulders, elbows, wrists, hips, knees, ankles, pelvis, thorax, upper neck, head top), the corresponding visibility of these joints, 3D torso orientations, head orientations, and activity labels. 
This dataset is split into the train/valid/test sets of size 25863/2958/11701.

\paragraph{\bf Look Into Person (LIP) dataset.
} 
Look Into Person (LIP) \cite{liang2018look} is another large-scale and challenging dataset for human parsing and pose estimation. It contains 50,462 human images (19,081 full-body images, 13,672 upper-body images, 403 lower-body images, 3,386 head-missed images, 2,778 back-view images, and 21,028 images with occlusions) with challenging view, severe occlusions, various postures, varying clothing, and low resolution. This dataset is divided into 30,462 training images, 10,000 validation images and 10,000 test images. Each person is annotated with 16 joints (i.e., shoulders, elbows, wrists, hips, knees, ankles, head, neck, spine, and pelvis).

\begin{table*}[htbp]
    \centering
    \setlength{\tabcolsep}{0.58cm} 
    \caption{Single person pose estimation datasets.}
    \begin{tabular}{l || c | c | c}
         \toprule
         \bf Dataset & \bf Training/Val/Test & \bf Parts &\bf \#Joints  \\
         \midrule 
         BBC Pose \cite{charles2014automatic}   
         & 10 videos/5 videos/5 videos       & Upper body 
         & 7  \\
         Extended BBC Pose \cite{charles2014automatic}
         & 82 videos/5 videos/5 videos       & Upper body
         & 7  \\
         Short BBC Pose \cite{charles2014automatic}
         & -/-/5 videos                      & Upper body
         & 7  \\
         Charlearn \cite{escalera2013multi}   
         & 393 videos/287 videos/276 videos  & Full body
         & 20 \\
         PiW \cite{cherian2014mixing}
         & total 30 videos                   & Upper body
         & 8  \\
         FLIC  \cite{sapp2013modec}
         & 3,987 images/-/1,016 images       & Upper body  
         & 10 \\
         FLIC-full  \cite{sapp2013modec}
         &  total 20928 images               & Upper body  
         & 10 \\
         LSP  \cite{johnson2010clustered}
         & 1,000 images/-/1,000 images       & Full body   
         & 14 \\
         Extended LSP   \cite{johnson2010clustered}
         & 11,000 images/-/1,000 images       & Full body   
         & 14 \\
         MHP    \cite{andriluka20142d}
         & 25,863 images/2,958 images/11,701 images  & Full body   
         & 16 \\
         LIP     \cite{liang2018look}      
         & 30,462 images/10,000 images/10,000 images & Full body   
         & 16 \\
         \bottomrule
    \end{tabular}
    \label{tab:dataset}
\end{table*}

\section{Performance Evaluation Metrics}
\label{metrics}
This section focuses on the model evaluation metrics
that differentiate the methods with some numerical index
regarding their generalization performance.

\paragraph{\bf Percentage of Correct Parts (PCP).
}
\begin{figure}
    \centering
    \resizebox{1\linewidth}{!}{%
        \includegraphics{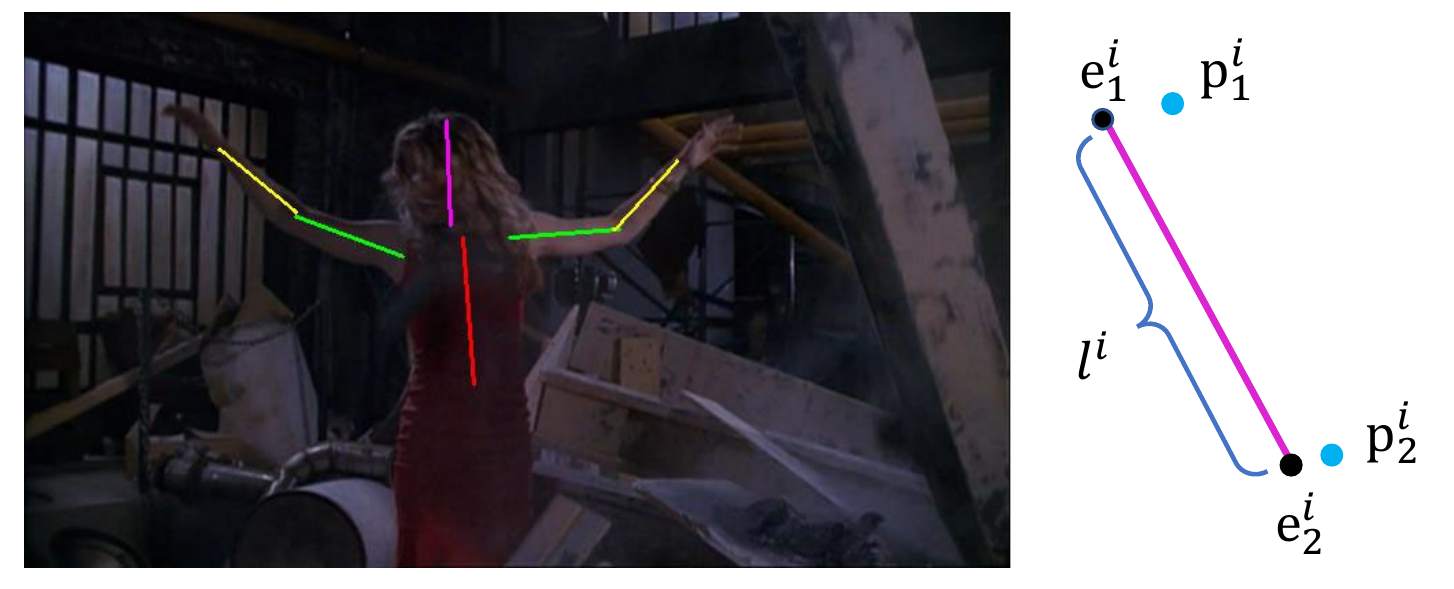}
    }%
    \caption{The stickmen annotation. $e_{1}^{i}$ and $e_{2}^{i}$ are the two endpoints of limb $i$ and $l^{i}$ is the limb length, $p_{1}^{i}$ and $p_{2}^{i}$ are the corresponding predictions.}
    \label{fig:stickmen}
\end{figure}
Ferrari et al. \cite{ferrari2008progressive} defined a performance criterion
based on the stickmen annotation as shown in Fig. \ref{fig:stickmen}. 
In the definition of stickmen, each limb $i$ of person $n$ is simplified as a segment which has two endpoints $ \{ p_{i,1}^{n}, p_{i,2}^{n} \} $.
When evaluating the Percentage of Correct Parts (PCP), the distance $d_{i,1}^{n}$ between the predicted joint $p_{i,1}^{n}$ and the corresponding endpoint $e_{i,1}^{n}$ is normalized by the limb length $l_{i}^{n} = \| e_{i,1}^{n}-e_{i,2}^{n} \|_2$ formulated as:
\begin{equation}
    \label{eq:pcp_dist1}
    d_{i,1}^{n} = \frac{\| p_{i,1}^{n} - e_{i,1}^{i} \|_2}{ l_{i}^{n} }  \\
\end{equation}
Similarly, we can obtain another distance $d_{i,2}^{n}$:
\begin{equation}
    \label{eq:pcp_dist2}
    d_{i,2}^{n} = \frac{\| p_{i,2}^{n} - e_{i,2}^{i} \|_2}{ l_{i}^{n} }  \\
\end{equation}

The original PCP takes both endpoints into consideration.
It calculates the distances from both endpoints to their corresponding predictions and then compares the average of both distances against a pre-set threshold to determine whether this limb is detected correctly: %
\begin{align}
\label{eq:pcp_condition}
\delta(i,n) =
\left\{ 
    \begin{array}{lc}
     1,  &  \frac{d_{i,1}^{n} + d_{i,2}^{n} }{2} \leq r \\
     0,  &  \text{otherwise}
    \end{array}
    \right.
\end{align}
\noindent where $r$ is the threshold and its value usually takes $50\%$ of the limb length.

The original PCP for limb $i$ is defined as:
\begin{equation}
\label{eq:PCP}
\text{PCP}(i) = \sum_{n=1}^{N} \frac{\delta(i,n)}{N}
\end{equation}

In addition to the original PCP metric, Johnson et al. \cite{johnson2010clustered} and Pishchulin et al. \cite{pishchulin2012articulated} adopted a stricter $\delta(i,n)$ as:
\begin{align}
\label{eq:pcp_condition2}
\delta(i,n) =
\left\{ 
    \begin{array}{lc}
     1,  &  d_{i,1}^{n} \leq r  \text{ and }  d_{i,2}^{n} \leq r \\
     0,  &  \text{otherwise}
    \end{array}
    \right.
\end{align}

\paragraph{\bf Percent of Detected Joints (PDJ).
}
Since the two PCP metrics penalize shorter limbs,
they cannot reveal the actual performance of a HPE method.
Toshev et al. \cite{toshev2014deeppose} designed the Percent of Detected Joints (PDJ) which instead uses the torso diameter $t^{n}$ as the scale factor to normalize the distance between the prediction $p_{j}^{n}$ and the ground truth $g_{j}^{n}$ for joint $j$ of person $n$ as:
\begin{equation}
    \label{eq:pdj_normdist}
    d_{j}^{n} = \frac{\| p_{j}^{n} - g_{j}^{n} \|_2}{ t^{n} }  \\
\end{equation}
Like the PCP metrics, if the normalized distance $d_{j}^{n}$ is within a threshold $r$, the joint $j$ of person $n$ is considered detected. This can be formulated as:
\begin{align}
\label{eq:pdj_condition}
\delta(j,n) & =
\left\{ 
    \begin{array}{lc}
     1,  &  d_{j}^{n} \leq r  \\
     0,  &  \text{otherwise}
    \end{array}
    \right.
\end{align}
Thus, the PDJ for joint $j$ can be defined as:
\begin{equation}
    \label{eq:pdj}
    \text{PDJ}(j) = \sum_{n=1}^{N} \frac{\delta(j,n)}{N}
\end{equation}

\paragraph{\bf Percentage of Correct Keypoints (PCK).
}
When evaluating the predictions of a model, 
the PDJ metric is greatly influenced by the torso diameter, 
making it less robust and consistent. 
Hence, Andriluka et al. \cite{andriluka20142d} modified the PDJ metric
by replacing the torso diameter with the head segment length
$h^{n}$:
\begin{align}
    d_j^n &= \frac{\| p_{j}^{n} - g_{j}^{n} \|_2}{h^{n}}
\end{align}
The resulting metric is 
named PCKh, defined as:
\begin{align}
    \label{eq:pck}
    \text{PCKh}(j) &= \sum^{N}_{n=1} \frac{\delta(d_j^n<0.5)}{N}
\end{align}
Besides, they also computed the Area under the PCK Curve (AUC) with the threshold ranging from $0$ to $0.5$ for giving a more comprehensive evaluation.

\paragraph{\bf Other Metrics.
}
In addition to the above metrics, the researchers have also applied other metrics to assess the performance of a method from different perspectives %
For example, the number of parameters is usually utilized to measure the
model complexity and the FLoating-point OPerations (FLOPs) is adopted to evaluate the computing complexity \cite{zhang2019fast}. 
The real-time performance is another key aspect for practical applications, and
the Frames Per Second (FPS) is an often-used metric to evaluate the running speed at certain devices.

\section{Conclusions and Discussions}
This paper has reviewed a wide range of recent researches on HPE in the following four perspectives: 
data augmentation, model architecture, post-processing, and learning target.
We discuss the development of HPE methods %
at data augmentation, model architecture, pose-processing, and 
supervision representation.
Although great improvements have been made,
HPE still faces some fundamental challenges,
as detailed below.

\paragraph{\bf Transferability.
}
Deep learning based HPE methods rely heavily on labelled data 
with specific characteristics.
For example, the MHP dataset covers daily activities, while the LSP dataset focuses on the sports scene.
The model trained on one dataset may perform badly on another dataset \cite{chen2021transfer}.
Thus, the transferability (i.e., model generalization across different domains) is still an important unsolved problem.

\paragraph{\bf Severe occlusions.
}
Occlusion resulting from varying shooting angles and crowd scenes remains a challenging problem in HPE.
Some methods tackles this problem by learning explicit or implicit spatial models \cite{wei2016convolutional,wang2020graph}, but suffer from the notorious overfitting issue.
To deal with the occlusion problem efficiently and effectively, 
inferring the positions of occluded joints might give useful guidance information.

\paragraph{\bf Low-resolution images.
}
Low-resolution images are widely available in the wild
and present extra challenges for state-of-the-art methods
due to the lack of fine-grained appearance information.
Nonetheless, it is still largely under-studied despite its potential importance in practice.
There have been a couple of preliminary attempts on low-resolution challenge \cite{neumann2018tiny,wang2021lowres}
with more advanced models to be innovated in the future.

\paragraph{\bf Real-time performance.
}
Current researches mostly focus on designing deeper and more complicated HPE model to obtain higher accuracies.
In many real-world applications, however, performing fast inference is likely a more important consideration 
especially for resource-limited platforms with human-computer interaction involved \cite{zhang2019fast}.

\paragraph{\bf System integration and applications.
}
The HPE is a fundamental task in computer vision with wide application values such as social robotics, human body generation, human-object interaction, activity recognition, person re-identification.
Taking into account the integration of HPE in the whole system and conducting its research accordingly would be more focusing and purposing due to the presence of contextual knowledge and domain priors.

\bibliographystyle{spbasic}      %
\bibliography{references}   %

\end{document}